\documentclass[runningheads]{llncs}
\usepackage{graphicx}
\usepackage{amsmath}
\usepackage[ruled]{algorithm2e}
\usepackage{rotating}
\usepackage{subcaption}
\usepackage{amsfonts}
\usepackage{booktabs}
\usepackage[misc]{ifsym}

\usepackage{hyperref}

\SetKwProg{Fn}{Function}{}{}

\begin{document}
\title{Powershap: A Power-full Shapley Feature Selection Method}
%
%

\author{Jarne~Verhaeghe \Letter \inst{1}\orcidID{0000-0002-3322-150X} \and
Jeroen~Van~Der~Donckt\inst{1}\orcidID{0000-0002-9620-888X} \and
Femke~Ongenae\inst{1}\orcidID{0000-0003-2529-5477} \and
Sofie~Van~Hoecke\inst{1}\orcidID{0000-0002-7865-6793}}

\tocauthor{}

\authorrunning{J. Verhaeghe et al.}

%

\institute{
IDLab, Ghent University - imec, 9052 Zwijnaarde
\email{jarne.verhaeghe@ugent.be}\\
\url{http://predict.idlab.ugent.be/}
}

\tocauthor{Jarne~Verhaeghe [IDLab, imec- Ghent University, 9052 Zwijnaarde]
Jeroen~Van~Der~Donckt [IDLab, imec- Ghent University, 9052 Zwijnaarde]
Femke~Ongenae [IDLab, imec- Ghent University, 9052 Zwijnaarde] 
Sofie~Van~Hoecke [IDLab, imec- Ghent University, 9052 Zwijnaarde] }

\toctitle{Powershap: A Power-full Shapley Feature Selection Method}
\maketitle              
\begin{abstract}
Feature selection is a crucial step in developing robust and powerful machine learning models. Feature selection techniques can be divided into two categories: filter and wrapper methods. While wrapper methods commonly result in strong predictive performances, they suffer from a large computational complexity and therefore take a significant amount of time to complete, especially when dealing with high-dimensional feature sets. Alternatively, filter methods are considerably faster, but suffer from several other disadvantages, such as (i) requiring a threshold value, (ii) many filter methods not taking into account intercorrelation between features, and (iii) ignoring feature interactions with the model. To this end, we present \textit{powershap}, a novel wrapper feature selection method, which leverages statistical hypothesis testing and power calculations in combination with Shapley values for quick and intuitive feature selection. \textit{Powershap} is built on the core assumption that an informative feature will have a larger impact on the prediction compared to a known random feature. Benchmarks and simulations show that \textit{powershap} outperforms other filter methods with predictive performances on par with wrapper methods while being significantly faster, often even reaching half or a third of the execution time. As such, \textit{powershap} provides a competitive and quick algorithm that can be used by various models in different domains. Furthermore, \textit{powershap} is implemented as a plug-and-play and open-source \textit{sklearn} component, enabling easy integration in conventional data science pipelines. User experience is even further enhanced by also providing an automatic mode that automatically tunes the hyper-parameters of the \textit{powershap} algorithm, allowing to use the algorithm without any configuration needed.

\keywords{Feature selection \and Shap \and Benchmark \and Simulation \and Toolkit \and Python \and Open source}
\end{abstract}

\section{Introduction}
In many data mining and machine learning problems, the goal is to extract and discover knowledge from data. One of the challenges frequently faced in these problems is the high dimensionality and the unknown relevance of features~\cite{li_feature_2017}. Ignoring these challenges will more than often result in modeling obstacles, such as sparse data, overfitting, and the curse of dimensionality. Therefore, feature selection is frequently applied, among other techniques, to effectively reduce the feature dimensionality. The smaller subset of features has the potential to explain the problem better, reduce overfitting, alleviate the curse of dimensionality, and even facilitate interpretation. Furthermore, feature selection is known to increase model performance, increase computational efficiency, and increase the robustness of many models due to the dimensionality reduction~\cite{li_feature_2017}. 

In this work, we present a novel feature selection method, called \textit{powershap}, that is a faster and easy-to-use wrapper method. The feature selection is realized by using Shapley values, statistical tests, and power calculations.

First, in Section \ref{sec:related}, a short overview of the related work is given to show how \textit{powershap} improves upon all these methods. Subsequently, in Section \ref{sec:powershap}, the method and the design choices are explained as well as the resulting algorithm. Finally, the performance of \textit{powershap} is compared to other state-of-the-art methods in Section \ref{sec:experiments} and \ref{sec:experiments_results} using both simulation and open-source benchmark datasets and the results are discussed in Section \ref{sec:discussion}. Finally, the conclusions are summarized in Section \ref{sec:conclusion}.

\section{Related Work}\label{sec:related}
Feature selection approaches can be categorized into filter and wrapper methods. Filter methods select features by measuring the relevance of the feature using model-agnostic measures, such as statistical tests, information gain, distance, similarity, and consistency to the dependent variable (if available). 
These methods are model-independent as this category of feature selection does not rely on training machine learning models~\cite{jovic_review_2015}, resulting in a fast evaluation. 
However, the disadvantages of filter methods are that they frequently impose assumptions on the data, are limited to a single type of prediction, such as classification or regression, not all methods take inter-correlation between features into account, and often require a cut-off value or hyperparameter tuning~\cite{kumari_filter_2011}. 
Examples of these filter methods are rank, chi² test, f-test, correlation-based feature selection, Markov blanket filter, and Pearson correlation~\cite{jovic_review_2015}. \\
Wrapper methods measure the relevance of features using a specific evaluation procedure through training supervised models. Depending on the wrapper technique, models are trained on either subsets of the features or on the complete feature set. The trained models are then utilized to select the resulting feature subset, using the aforementioned performance metrics, or by ranking the inferred feature importances. In general, wrapper methods tend to provide smaller and more qualitative feature subsets than filter methods, as they take the interaction between the features, and between the model and the features, into account~\cite{colaco_review_2019}. A major drawback of wrapper methods is the considerable time complexity associated with the underlying search algorithm, or in the case of feature importance ranking the hyperparameter tuning. Examples of wrapper methods are forward, backward, genetic, or rank-based feature importance feature selection.\\
In the interpretable machine learning field, one of the emerging and proven techniques to explain model predictions is SHAP~\cite{NIPS2017_7062}. This technique aims at quantifying the impact of features on the output. To do so, SHAP uses a game-theory inspired additive feature-attribution method based on Shapley Regression Values~\cite{NIPS2017_7062}. This method is model-agnostic and implemented for various models, e.g., linear, kernel-based, deep learning, and tree-based models. Although SHAP suffers from shortcomings, such as its TreeExplainer providing non-zero Shapley values to noise features, it is technically strong and very popular~\cite{linardatos_explainable_2020}. \\
The strength of the SHAP algorithm facilitates the development of new feature selection methods using Shapley values. A simple implementation would be a rank-based feature selection, which ranks the different features based on their Shapley values on which a rank cut-off value determines the final feature set. However, there are more advanced methods available. One of these more advanced techniques is borutashap~\cite{eoghan_keany_2020_4247618}. Borutashap is based on the Boruta algorithm that makes use of shadow features, i.e. features with randomly shuffled values. Boruta is built on the idea that a feature is only useful if it is doing better than the best-performing shuffled feature. To do so, Boruta compares the feature importance of the best shadow feature to all other features, selecting only features with larger feature importance than the highest shadow feature importance. Statistical interpretation is realized by repeating this algorithm for several iterations, resulting in a binomial distribution which can be used for p-value cut-off selection~\cite{boruta_base}. Borutashap improves on the underlying Boruta algorithm by using Shapley values and an optimized version of the shap TreeExplainer~\cite{eoghan_keany_2020_4247618}. As such, implementations of Borutashap are limited to tree-based models only.\\
Another shap-based feature selection method using statistics is shapicant~\cite{calzolari_manuel-calzolarishapicant_2022}. This feature selection method is inspired by the permutation-importance method, which first trains a model on the true dataset, and afterward, it shuffles the labels and retrains the model on the shuffled dataset. This process is repeated for a set amount of iterations, from which the average feature importances of both models are compared. If for a specific feature, the feature importance of the true dataset model is consistently larger than the importance of the shuffled dataset model, that feature is considered informative. Using a non-parametric estimation it is possible to assign a p-value to determine a wanted cut-off value~\cite{altmann_permutation_2010}. Shapicant improves on this underlying algorithm by using Shapley values. Specifically, it uses both the mean of the negative and positive Shapley values instead of Gini importances, which are only positive and frequently used for tree-based model importances. Furthermore, shapicant uses out-of-sample feature importances for more accurate estimations and an improved non-parametric estimation formula~\cite{calzolari_manuel-calzolarishapicant_2022}. \\
\textit{Powershap} draws inspiration from the non-parametric estimation of shapicant and the random feature usage in borutashap and improves upon all these state-of-the-art filter and wrapper algorithms resulting in at least comparable performances while being significantly faster.

\section{Powershap}\label{sec:powershap}
\textit{Powershap} builds upon the idea that a known random feature should have, on average, a lower impact on the predictions than an informative feature. To realize feature selection, the \textit{powershap} algorithm consists of two components: the \textit{Explain} component and the core \textit{powershap} component. First, in the \textit{Explain} part, multiple models are trained using different random seeds, on different subsets of the data. Each of these subsets is comprised of all the original features together with one random feature. Once the models are trained, the average impact of the features (including the random feature) is explained using Shapley values on an out-of-sample dataset. Then, in the core \textit{powershap} component, the impacts of the original features are statistically compared to the random feature, enabling the selection of all informative features. 

\subsection{Powershap Algorithm}
In the \textit{Explain} component, a single known random uniform (RandomUniform) feature is added to the feature set for training a machine learning model. Unlike the Boruta algorithm, where all features are duplicated and shuffled, only a single random feature is added. In some models, such as neural networks, duplicating the complete feature set increases the scale and thereby increases the time complexity drastically. Using the Shapley values on an out-of-sample subset of the data allows for quantifying the impact on the output for each feature. The Shapley values are evaluated on unseen data to assess the true unbiased impact~\cite{breiman_random_2001}. As a final step, the absolute value of all the Shapley values is taken and then averaged ($\mu$) to get the total average impact of each feature. Compared to shapicant, only a single mean value is used here, resulting in easier statistical comparisons. Furthermore, by utilizing the absolute Shapley values, the positive values and the negative values are added to the total impact, which could result in a different distribution compared to the Gini importance. This procedure is then repeated for $I$ iterations, where every iteration retrains the model with a different random feature and uses a different subset of the data to quantify the Shapley values, resulting in an empirical distribution of average impacts that will further be used for the statistical comparison. In the codebase, the procedure explained above is referred to as the \textit{Explain} function. The pseudocode of the \textit{Explain} function is shown in Algorithm~\ref{alg:explain}.

\begin{algorithm}
    \DontPrintSemicolon
    
    \Fn{Explain($I\leftarrow$ Iterations, $M\leftarrow$ Model,  $\mathbf{D}^{n \times m}\leftarrow$ Data, $rs\leftarrow$ Random seed)}{
    $\mathbf{powershap}_{values}\leftarrow $size $[I,m+1]$\;
    \For{$i\leftarrow 1,2,\dots,I$}{
        $RS\leftarrow i+rs$\;
        $D_{random}^{n}\leftarrow$ RandomUniform$(RS)$ $\in [-1,1]$ size $n$\;
        $\mathbf{D}^{n\times m+1} \leftarrow \mathbf{D}^{n \times m} \cup D_{random}^{n}$\;
        $\mathbf{D}_{train}^{0.8n \times m+1},\mathbf{D}_{val}^{0.2n \times m+1}\leftarrow$ split $\mathbf{D}$\;
        $M \leftarrow$ Fit $M(\mathbf{D}_{train})$\;
        $\mathbf{S}_{values} \leftarrow$ SHAP$(M$, $\mathbf{D}_{val})$\;
        $\mathbf{S}_{values} \leftarrow |\mathbf{S}_{values}|$\;
        \For{$j\leftarrow 1,2,\dots,m+1$}{
            $\mathbf{powershap}_{values}[i][j]\leftarrow \mu(\mathbf{S}_{values}[\dots][j])$\;
        }
    }
    \textbf{return} $\mathbf{powershap}_{values}$
    }
    \caption{Powershap Explain algorithm}\label{alg:explain}
\end{algorithm}

Given the average impact of each feature for each iteration, it is then possible to compare it to the impact of the random feature in the core powershap component. This comparison is quantified using the percentile formula shown in Equation~\ref{eqn:percentile} where $\mathbf{s}$ depicts an array of average Shapley values for a single feature with the same length as the number of iterations, while $x$ represents a single value, and $\mathbb{I}$ represents the indicator function. This formula calculates the fraction of iterations where $x$ was higher than the average shap-value of that iteration and can therefore be interpreted as the p-value. 
\begin{equation}\label{eqn:percentile}
    Percentile(\mathbf{s},x)=\sum_{i}^{n}\frac{\mathbb{I}(x>s_i)}{n}
\end{equation}
Note that this formula provides smaller p-values than what should be observed, the correct empirical formula is $(1+\sum_{i}^{n} \mathbb{I}(x>s_i))/(n+1)$ as explained by North et al.~\cite{north_note_2002}. This issue of smaller p-values mainly persists for lower number of iterations. However, \textit{powershap} implements Equation~\ref{eqn:percentile} as this anticonservative estimation of the p-value is desired behavior for the automatic mode (see Section~\ref{sec:automatic}). This formula enables setting a static cut-off value for the p-value instead of a varying cut-off value and results in fewer required iterations, while still providing correct results. This will be further explained at the end of Section~\ref{sec:automatic}. \\
As the hypothesis states that the impact of the random feature should be on average lower than any informative feature, all impacts of the random feature are again averaged, resulting in a single value that can be used in the percentile function. This results in a p-value for every original feature. This p-value represents the fraction of cases where the feature is less important, on average than a random feature. Given the hypothesis and these p-value calculations, a heuristic implementation of a one-sample one-tailed student-t smaller statistic test can be done, where the null hypothesis states that the random feature ($H_1$-distribution) is not more important than the tested feature ($H_0$-distribution)~\cite{lomax_introduction_2007}. Therefore, the positive class in this statistical test represents a true null hypothesis. This heuristic implementation does not assume a distribution on the tested feature impact scores, in contrast to a standard student-t statistic test where a standard Gaussian distribution is assumed. Then, given a threshold p-value $\alpha$, it is possible to find and output the set of informative features. The pseudocode of Algorithm~\ref{alg:base} details how the core \textit{powershap} feature selection method is realized.

\begin{algorithm}
    \DontPrintSemicolon
    
    \Fn{Powershap ($I\leftarrow$ Iterations, $M\leftarrow$ Model, $\mathbf{F}_{set} \leftarrow {F_{1},\dots,F_{m}}$, $\mathbf{D}\leftarrow$ Data size $[n,m]$, $\alpha \leftarrow$ required p-value)}{
    
    $\mathbf{powershap}_{values} \leftarrow$\textbf{Explain}($I$, $M$, $\mathbf{D}$)\;
    $S_{random}\leftarrow\mu(\mathbf{powershap}_{values}[...][m+1])$\;
    $\mathbf{P}^{m} \leftarrow$ initialize\;
    \For{$j\leftarrow 1,2,\dots,m$}{
        $\mathbf{P}[j]\leftarrow Percentile(\mathbf{powershap}_{values}[...][j]$, $S_{random})$\;
    }
    \textbf{return} $\{F_i\ |\ \forall$ $i$ : $\mathbf{P}[i]<\alpha\}$ 
    }
    \caption{Powershap core algorithm}\label{alg:base}
\end{algorithm}

\subsection{Automatic Mode}\label{sec:automatic}
\begin{figure}
     \centering
     \includegraphics[width=\textwidth]{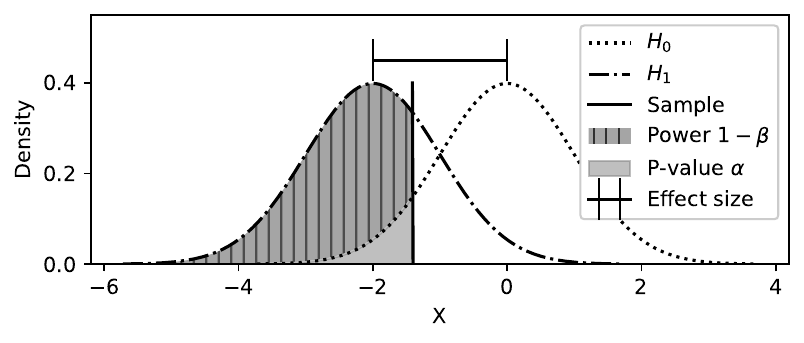}
     \caption{Visualization of p-value, effect size, and power for a standard t-test.}\label{power_expl}
\end{figure}

Running the \textit{powershap} algorithm consisting of the \textit{explain} and the \textit{core} components, requires setting two hyperparameters: $\alpha$ the p-value threshold and $I$ the number of iterations. When hyperparameter tuning, one should make a trade-off between runtime and quality. On the one hand, there should be enough iterations to avoid false negatives for a given $\alpha$, especially with the anticonservative p-values. On the other hand, adding iterations increases the time complexity. To avoid the need for users to manually optimize these two hyperparameters, \textit{powershap} also has an automatic mode. This automatic mode, automatically determines and optimizes the iteration hyperparameter $I$ using statistical power calculation for $\alpha$, hence the name \textit{powershap}.\\
The statistical power of a test is $1-\beta$, where $\beta$ is the probability of false negatives. In this case, a false negative is a non-informative feature flagged as an informative one. If a statistical test of a tested sample outputs a p-value $\alpha$, this represents the chance that the tested sample could be flagged as \textit{significant} by chance given the current data. This is calculated using Equation~\ref{eqn:alpha}. If the data in the statistical test is small, it is possible to have a very low $\alpha$ but a large $\beta$, resulting in an output that cannot be trusted. Therefore, for a given $\alpha$, the associated power should be as close to 1 as possible to avoid any false negatives. The power of a statistical test can be calculated using the cumulative distribution function $F$ of the underlying tested distribution $H_1$ using Equation~\ref{eqn:power}. Figure~\ref{power_expl} explains this visually. In the current context, $H_0$ could represent the random feature impact distribution and $H_1$ the tested feature impact distribution.

\begin{equation}\label{eqn:alpha}
    \alpha(x)=F_{H_0}(x)
\end{equation}
\begin{equation}\label{eqn:power}
    Power(\alpha)=F_{H_1}\left(F^{-1}_{H_0}(\alpha)\right)
\end{equation}

The power calculations require the cumulative distribution function $F$. However, the underlying distributions of the calculated feature impacts are unknown. In addition, calculating $F$ heuristically does not enable calculating the required iteration hyperparameter, which is the goal of the automatic mode. \textit{Powershap} circumvents this by mapping the underlying distributions to two standard student-t distributions as visualized in Figure \ref{power_expl}. It first calculates the pooled standard deviation, using Equation~\ref{eqn:pooledstd}, by averaging the standard deviations $\sigma$ of both distributions. It then calculates the distance $d$ between these two distributions, also called the effect size, in terms of this pooled standard deviation using the Cohen's d effect size as detailed in  Equation~\ref{eqn:effectsize}~\cite{lomax_introduction_2007}. Now, it is possible to define two standard student-t distributions with distance $\sqrt{I}\cdot d$ apart and $I-1$ degrees of freedom, where $I$ is the amount of \textit{powershap} iterations. The standard central student-t $F_{CT}$ and non-central student-t $F_{NCT}$ cumulative distribution functions are then used to calculate the power of the statistical test according to Equation~\ref{eqn:ttestpower}. This equation can in turn be used in a heuristic algorithm to solve for $I$. \textit{Powershap} uses the \textit{solve\_power} implementation of \textit{statsmodels} to determine the required $I$ from the TTestPower equation using brentq expansion for a provided required power~\cite{seabold2010statsmodels}. The \textit{powershap} pseudocode for the calculation of the effect size, power, and required iterations is shown in Algorithm~\ref{alg:analysis}.
\begin{equation}\label{eqn:pooledstd}
    PooledStd(\mathbf{s}_1,\mathbf{s}_2)=\frac{\sqrt{(\sigma^2 (\mathbf{s}_1)+\sigma^2 (\mathbf{s}_{2}))}}{2}
\end{equation}
\begin{equation}\label{eqn:effectsize}
    EffectSize(\mathbf{s}_1,\mathbf{s}_2)=\frac{\mu(\mathbf{s}_1)-\mu(\mathbf{s}_2)}{PooledStd(\mathbf{s}_1,\mathbf{s}_2)}
\end{equation}
\begin{equation}\label{eqn:ttestpower}
    TTestPower(\alpha,I,d_{calc})=F_{NCT}\left(F^{-1}_{CT}(\alpha,k=I-1),k=n-1,d=\sqrt{I}\cdot d_{calc}\right)
\end{equation}

\begin{algorithm}
    \DontPrintSemicolon
    \Fn{Analysis($\alpha \leftarrow$ required p-value, $\beta \leftarrow$ required power, $\mathbf{powershap}_{values}$)}{
    $\mathbf{S}_{random}\leftarrow \mathbf{powershap}_{values}[...][m+1]$\;
    $\mathbf{P}\leftarrow$ size $[m]$\;
    $\mathbf{N}_{required}\leftarrow$ size $[m]$\;
    \For{$j\leftarrow 1,2,\dots,m$}{
        $\mathbf{S}_i \leftarrow \mathbf{powershap}_{values}[...][j]$\;
        $\mathbf{P}[j]\leftarrow Percentile(\mathbf{S}_i$, $\mu(S_{random}))$\;
        effectsize $\leftarrow$ EffectSize($\mathbf{S}_i$, $\mathbf{S}_{random}$)\;
        $\mathbf{N}_{required}\leftarrow$ \textbf{SolveTTestPower}(effectsize, $\alpha, \beta$)\;
    }
    \textbf{return} $\mathbf{P}$, $\mathbf{N}_{required}$
    }
    \caption{Powershap analysis function}\label{alg:analysis}
\end{algorithm}
With the calculated required amount of iterations $n$, the automatic \textit{powershap} algorithm can be executed. The pseudocode to enable the automatic mode is shown in Algorithm~\ref{alg:automatic}. As can be seen, this is an expansion of the core algorithm (see Algorithm~\ref{alg:base}) and starts with an initial ten iterations to calculate the initial p-value, effect sizes, power, and required iterations for all features. Then, it searches for the largest required number of iterations $I_{max}$ of all tested features having a p-value below the threshold $\alpha$. If $I_{max}$ exceeds the already performed number of iterations $I_{old}$, automatic mode continues \textit{powershap} for the extra required iterations. This process is repeated until the performed iterations exceed the required iterations. For optimization, when the extra required iterations ($I_{max}-I_{old})$ exceed ten iterations, the automatic mode first adds ten iterations and then re-evaluates the required iterations because the required iterations are influenced by the already performed iterations. Furthermore, it is also possible to provide a stopping criterion on the re-execution of \textit{powershap} to avoid an infinite calculation. As a result the time complexity of the algorithm is linear in terms of the underlying model and shap explainer and can be formulated as $O(p[M_{n+1}+S(M_{n+1}])$, with $n$ the amount of features, $p$ the number of powershap iterations, $S$ the shap explainer time, and $M_x$ the model fit time for $x$ features.
For the automatic mode, by default, $\alpha$ is set to $0.01$ while the required power is set to $0.99$. This results in only selecting features that are more important than the random feature for all iterations. Furthermore, this also compensates for the anticonservative p-value and avoids as many false negatives as possible. Realizing the same desired behavior with the more accurate p-value estimation would require a varying $\alpha$ of $1/n$, complicating the power calculations and increasing the likelihood of false negatives.
The resulting powershap algorithm is implemented in Python as an open-source plug-and-play \textit{sklearn} compatible component to enables direct usage in conventional machine learning pipelines~\cite{scikit-learn}. The codebase  \footnote{The code, documentation, and more benchmarks can be found using the following link: \href{https://github.com/predict-idlab/PowerSHAP}{https://github.com/predict-idlab/PowerSHAP}} already supports a wide variety of models, such as linear, tree-based, and even deep learning models. To assure the quality and correctness of the implementation, we tested the functionality using unit testing.

\begin{algorithm}
    \DontPrintSemicolon
    
    \Fn{Powershap ($M\leftarrow$ Model, $\mathbf{F}_{set} \leftarrow {F_{1},\dots,F_{m}}$, $\mathbf{D}^{n \times m}\leftarrow$ Data, $\alpha \leftarrow$ required p-value, $\beta \leftarrow$ required power)}{

    $\mathbf{powershap}_{values} \leftarrow$\textbf{Explain}($I\leftarrow 10$, $M$, $\mathbf{D}$, $rs\leftarrow 0$)\;
    
    $\mathbf{P},\mathbf{N}_{required}\leftarrow$\textbf{Analysis}($\alpha$, $\beta$, $\mathbf{powershap}_{values}$)\;
    
    $I_{max}\leftarrow$ ceil($\mathbf{N}_{required}$[MaxArg($\mathbf{P}<\alpha$)])\;
    $I_{old}\leftarrow 10$\;
    
    \While{$I_{max}>I_{old}$}{
        \eIf{$I_{max}-I_{old}>10$}{
            $\mathbf{auto}_{values} \leftarrow$\textbf{Explain}($I\leftarrow 10$, $M$, $\mathbf{D}$, $rs\leftarrow 0$)\;
            $I_{old}\leftarrow I_{old}+10$\;
        }
        {
            $\mathbf{auto}_{values} \leftarrow$\textbf{Explain}($I\leftarrow I_{max}-I_{old}$, $M$, $\mathbf{D}$, $rs\leftarrow 0$)\;
            $I_{old}\leftarrow I_{max}$\;
        
        }
        $\mathbf{powershap}_{values} \leftarrow \mathbf{powershap}_{values} \cup \mathbf{auto}_{values}$\;
        
        $\mathbf{P},\mathbf{N}_{required}\leftarrow$\textbf{Analysis}($\alpha$, $\beta$, $\mathbf{powershap}_{values}$)\;
        
        $I_{max}\leftarrow$ ceil($Max(\mathbf{N_{required}}[i, \forall i : \mathbf{P}[i]<\alpha]))$\;
        
    }
    
    \textbf{return} $[F_i , \forall$ $i$ : $\mathbf{P}[i]<\alpha]$ 

    }
    \caption{Automatic Powershap algorithm version}\label{alg:automatic}
\end{algorithm}

\section{Experiments}\label{sec:experiments}
\subsection{Feature Selection Methods}
To facilitate a comparison with other feature selection techniques, we benchmark \textit{powershap} together with other frequently used techniques on both synthetic and real-world datasets. In particular, \textit{powershap} is compared with both filter and wrapper methods, and state-of-the-art shap-based wrapper methods. To provide a fair comparison, all methods, including \textit{powershap}, were used in their default out-of-the-box mode without tuning. For \textit{powershap}, this default mode is the automatic mode.
Concerning filter methods, two methods were chosen: the chi-squared and f-test feature selection from the \textit{sklearn}-library~\cite{scikit-learn}. The chi-squared test measures the dependence between a feature and the classification outcome and assigns a low p-value to features that are not independent of the outcome. As the chi-squared test only works with positive values, the values are shifted in all chi-squared experiments such that all values are positive. This has no effect on tree-estimators as they are invariant to data scaling~\cite{lomax_introduction_2007}. The F-test in \textit{sklearn} is a univariate test that calculates the F-score and p-values on the predictions of a univariate fitted linear regressor with the target~\cite{scikit-learn}. Both filter methods provide p-values that are set to the same threshold as \textit{powershap}. 
As wrapper feature selection method, forward feature selection was chosen. This method is a greedy algorithm that starts with an empty set of features and trains a model with each feature separately. In every iteration, forward feature selection then adds the best feature according to a specified metric, often evaluated in cross-validation, until the metric stops improving. This is generally considered a strong method but has a very large time complexity~\cite{jovic_review_2015}.
\textit{Powershap} is also compared to shapicant~\cite{calzolari_manuel-calzolarishapicant_2022} and borutashap~\cite{eoghan_keany_2020_4247618}, two SHAP-based feature selection methods.
The default machine learning model used for all datasets and all feature selection methods, including \textit{powershap}, is a CatBoost gradient boosting tree-based estimator using 250 estimators with the overfitting detector enabled. For classification, the CatBoost model uses adjusted class weights to compensate for any potential class imbalance. The Catboost estimator often results in strong predictive performances out-of-the-box, without any hyper-parameter tuning, making it the perfect candidate for benchmarking and comparison~\cite{prokhorenkova_catboost_2019}. All experiments are performed on a laptop with a Intel(R) Core(TM) i7-9850H CPU at 2.60GHz  processor and 16 GB RAM running at 2667 MHz, with background processes to a minimum.

\subsection{Simulation Dataset}
The methods are first tested on a simulated dataset to assess their ability to discern noise features from informative features. The used simulation dataset is created using the \texttt{make\_classification} function of \textit{sklearn}. This function creates a classification dataset, however, exactly the same can be done for obtaining a regression dataset (by using \texttt{make\_regression}). The simulations are run using 20, 100, 250, and 500 total features to understand the performance on varying dimensions of feature sets. The ratio of informative features is varied as  10\%, 33\%, 50\%, and 90\% of the total feature set, allowing for assessing the quality of the selected features in terms of this ratio. The resulting simulation datasets each contain 5000 samples. Each simulation experiment was repeated five times with different random seeds. The number of redundant features, which are linear combinations of informative features, and the number of duplicate features were set to zero. Redundant features and duplicate features reduce the performance of models, but they cannot be discerned from true informative features as they are inherently informative. Therefore they are not included in the simulation dataset as the goal of \textit{powershap} is to find informative features.  The \textit{powershap} method is compared to shapicant, chi², borutashap, and the f-test for feature selection on this simulation dataset. Due to time complexity constraints, forward feature selection was not included in the simulation benchmarking. 

\subsection{Benchmark Datasets}

\begin{table}
\begin{center}
\caption{Properties of all datasets}
\label{table:dataset_properties}
\setlength{\tabcolsep}{2pt}
\begin{tabular}{cccccc}\toprule

Dataset & Type & Source & \# features & train size & test size \\\midrule
Madelon & Classification & OpenML & 500& 1950 & 650 \\
Gina priori & Classification & OpenML & 784 & 2601 & 867  \\
Scene & Classification & OpenML & 294 & 1805 & 867  \\
CT location & Regression & UCI &  384 & 41347 & 12153  \\
Appliances & Regression & UCI & 30 & 14801 & 4934  \\\bottomrule
\end{tabular}
\end{center}
\end{table}

In addition to the simulation benchmark, the different methods are also evaluated on five publicly available datasets, i.e. three classification datasets: the Madelon~\cite{vanschoren_openml_nodate}, the Gina priori~\cite{vanschoren_openml_nodate-1}, and the Scene dataset~\cite{vanschoren_openml_nodate-2}, and two regression datasets: CT location~\cite{Dua:2019} and Appliances~\cite{Dua:2019}. The details of these datasets are shown in Table~\ref{table:dataset_properties}. The Scene dataset is a multi-label dataset, however, a multi-label problem can always be reduced to a one-vs-all classification problem. Therefore only the label ``Urban" was chosen here to assess binary classification performance.\\
Almost all of these datasets have a large feature set, ideal for benchmarking feature selection methods. The datasets are split into a training and test set using a $75/25$ split. All methods are evaluated using both 10-fold cross-validation on the training set and 1000 bootstraps on the test set to assess the robustness of the performance. The test set is utilized to assess generalization beyond the validation set as wrapper methods tend to slightly overfit their validation set~\cite{jovic_review_2015}, while the training set is used for feature selection. The forward feature selection method was performed with 5-fold cross-validation and not 10-fold cross-validation due to the high time complexity. A validation set of 20\% of the training set is used for shapicant, using the same validation size as \textit{powershap} in Algorithm~\ref{alg:explain}. The models are evaluated with the AUC metric for classification datasets and with the $R^2$ metric for regression datasets.

\section{Results}\label{sec:experiments_results}
\subsection{Simulation Dataset}

\begin{sidewaysfigure}
    \includegraphics[width=\textwidth]{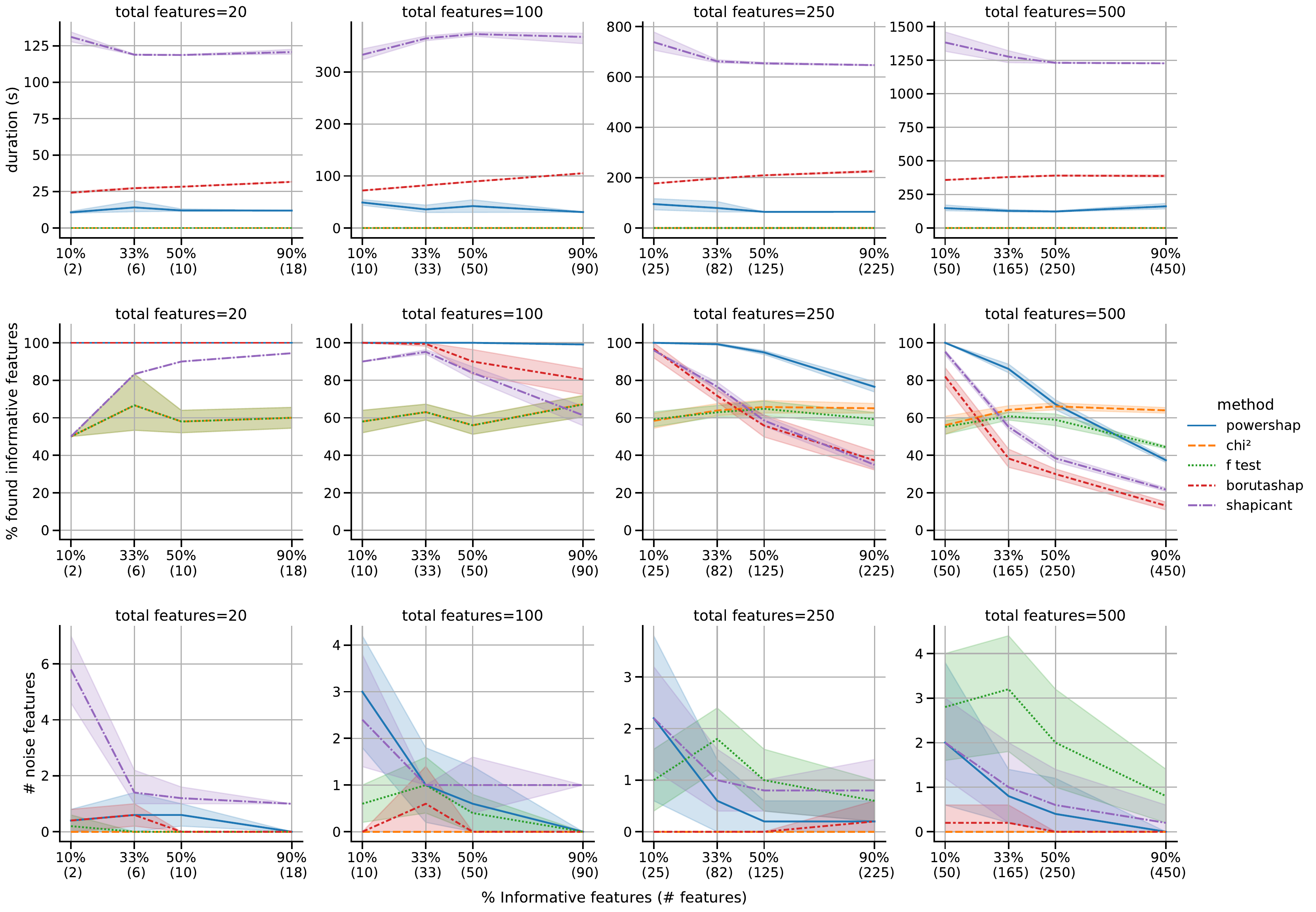}
    \caption{Simulation benchmark results using the \textit{make\_classification} \textit{sklearn} function for 5000 samples with five different \textit{make\_classification} random seeds.}
    \label{fig:total_sim_performance}
\end{sidewaysfigure}

The results of the simulation benchmarking are shown in Figure~\ref{fig:total_sim_performance}. Each row of subfigures shows the duration, the percentage of informative features found, and the number of selected noise features.  These measures are shown for each feature selection method for varying feature set dimensions and varying amounts of informative features. As can be seen, the shapicant method is the slowest wrapper method while \textit{powershap} is, without doubt, the fastest wrapper method. The filter methods are substantially faster than any of the wrapper methods, as they do not train models. Furthermore, \textit{powershap} finds all informative features with a limited amount of outputted noise features up to the case with 250 total features with 50\% (125) informative features, outperforming every other method. This can be explained by the model underfitting the data. Even with higher dimensional feature sets, \textit{powershap} finds more informative features than the other methods. Interestingly, most methods do not output many noise features, except for shapicant in the experiment with 20 total and 10\% informative features. 

\subsection{Benchmark Datasets}

Table~\ref{table:dataset_results_time} shows the duration of the feature selection methods and the size of the selected feature sets for each method on the different open-source datasets. Chi² does not apply to regression problems and is therefore not included in the results of the CT location and Appliances datasets. The table shows that \textit{powershap} is again the fastest wrapper method, while the number of selected features is in line with the other methods. The filter methods tend to output more features, while forward feature selection outputs a more conservative set of features.

\begin{table}
\begin{center}
\caption{Benchmarks results for duration and selected features. "default" indicates no feature selection or all features.}
\setlength{\tabcolsep}{4pt}
\label{table:dataset_results_time}
\begin{tabular}{cccccccc}\toprule
 \multicolumn{8}{c}{\textbf{duration (s)}} \\\cmidrule(lr){1-8}
 \textbf{Dataset} & \multicolumn{1}{c}{powershap }  & \multicolumn{1}{c}{borutashap } & \multicolumn{1}{c}{shapicant} & \multicolumn{1}{c}{forward } & \multicolumn{1}{c}{chi²} & \multicolumn{1}{c}{f test}& \multicolumn{1}{c}{default}\\\midrule
\textbf{Madelon} & 132s & 186s & 632s & 10483s & $<1s$ & $<1s$ & N/A \\
\textbf{Gina priori} & 184s & 299s & 812s & 68845s & $<1s$ & $<1s$ & N/A \\
\textbf{Scene} & 115s & 220s & 749s & 12496s & $<1s$ & $<1s$ & N/A \\
\textbf{CT location} & 459s & 543s & 1553s & 56879s &  N/A  & $<1s$ & N/A \\
\textbf{Appliances} & 34s & 48s &134s & 1913s &  N/A & $<1s$ & N/A \\\midrule
\multicolumn{8}{c}{\textbf{selected features}} \\\cmidrule(lr){1-8}

\textbf{Madelon} & 22 & 10 & 30 & 8 & 43 & 18 & 500\\
\textbf{Gina priori} & 105 & 37 & 106 & 26 & 328 & 405 & 784 \\
\textbf{Scene} & 36 & 14 & 56 & 15 & 93 & 220 & 294 \\
\textbf{CT location} & 123 & 162 & 74 & 75 & N/A & 350 & 384 \\
\textbf{Appliances} & 24 & 24 & 10 & 13 & N/A & 20 & 30\\\bottomrule
\end{tabular}
\end{center}
\end{table}

\begin{figure}
     \centering
     \begin{subfigure}{\textwidth}
         \centering
         \includegraphics[width=\textwidth]{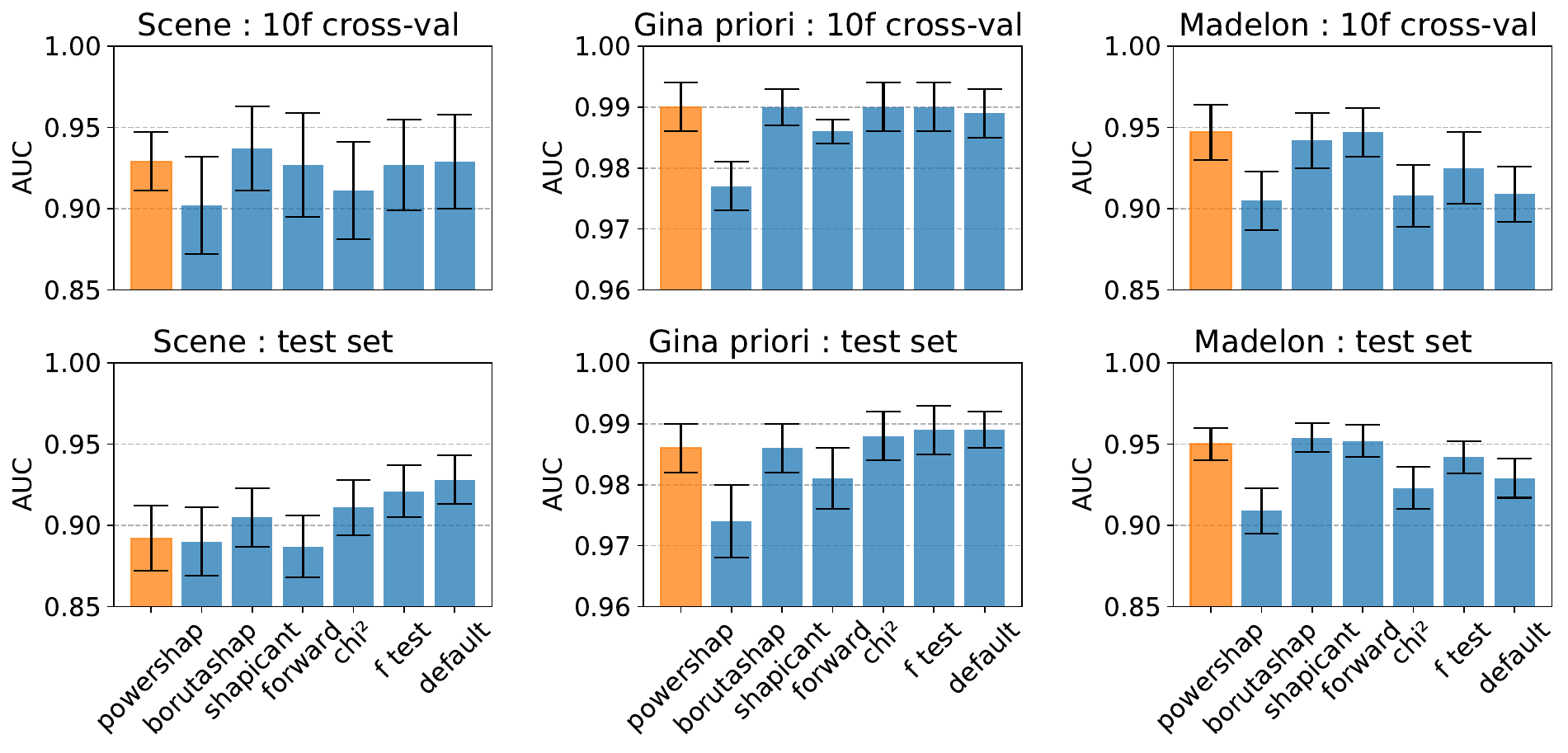}
         \caption{Classification benchmark dataset performances.}
         \label{fig:classification_benchmark}
     \end{subfigure}
     \begin{subfigure}{\textwidth}
         \centering
         \includegraphics[width=0.66\textwidth]{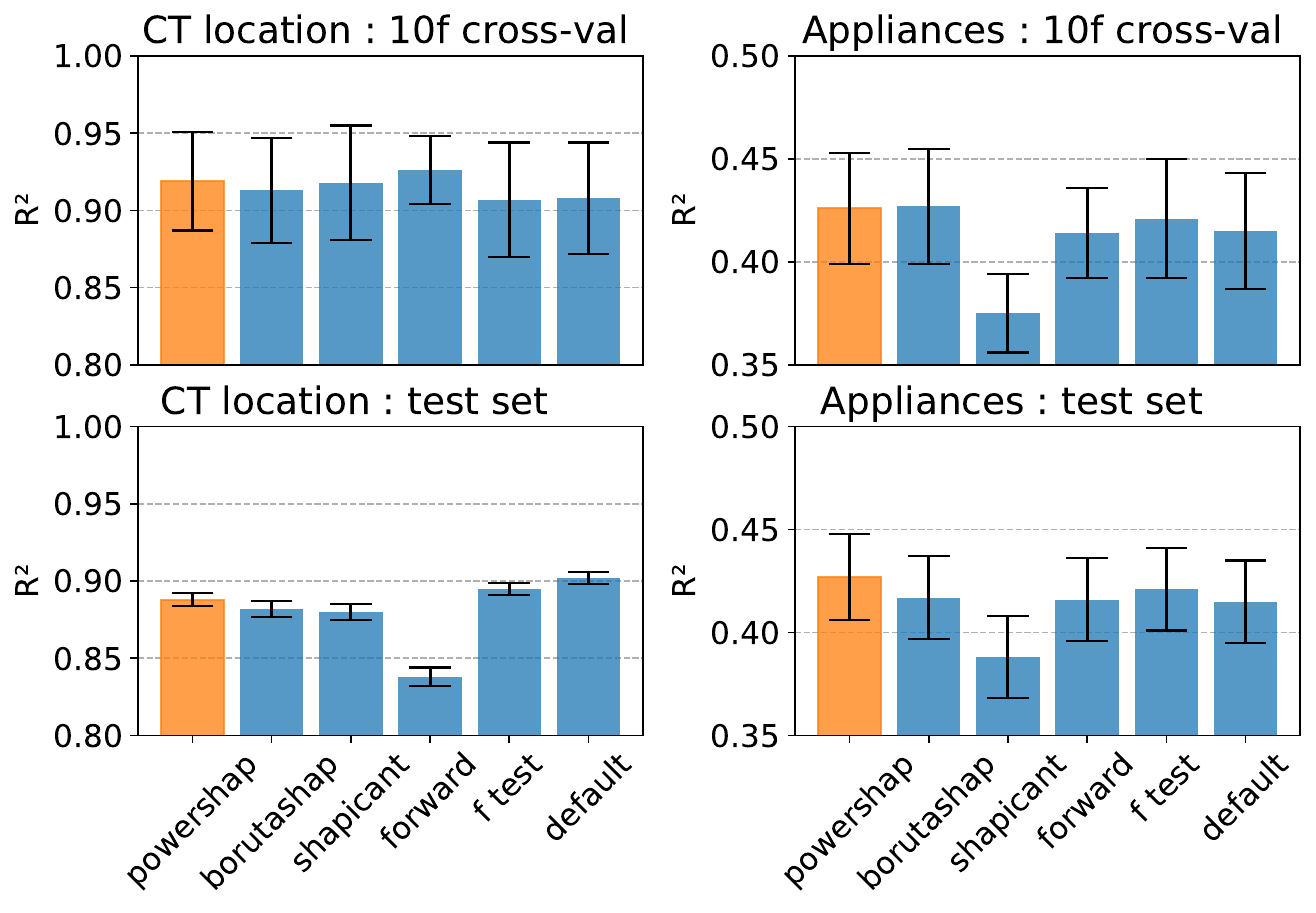}
         \caption{Regression benchmark dataset performances.}
         \label{fig:regression_benchmark}
     \end{subfigure}
    \caption{Benchmark performances. The error bars represent the standard deviation. }
    \label{fig:benchmark_bar_performance}
\end{figure}

The performance of the selected feature sets for each classification benchmark dataset is shown in Figure~\ref{fig:classification_benchmark} and in Figure~\ref{fig:regression_benchmark} for the regression benchmarks. These figures show that \textit{powershap} provides a steady performance on all datasets, consistently achieving the best or equal performance on both the cross-validation and test sets. However, even in cases with equal performance, \textit{powershap} achieves these performances considerably quicker, especially compared to shapicant and forward feature selection. The CT location dataset performances show that forward feature selection tends to overfit on the cross-validation dataset while \textit{powershap} is more robust.

\section{Discussion}\label{sec:discussion}

For the above test results, we used the default automatic \textit{powershap} implementation. However, similar to many other feature selection methods, \textit{powershap} can be further optimized or tuned. One of these optimizations is the use of a \textit{convergence} mode to extract as many informative features as possible. In this mode, \textit{powershap} continues recursively in automatic mode where in every recursive iteration, \textit{powershap} re-executes but with any previously found and selected features excluded from the considered feature set. This process continues, until no more informative features can be found. The convergence mode is especially useful in use-cases with high dimensional feature sets or datasets with a large risk of underfitting as it reduces the feature set dimension each recursive iteration to facilitate finding new informative features. As a basic experiment on the simulation benchmark, using the convergence mode for 500 features and 90\% (450) informative features, the percentage of found features increases from around 38\% (170) to 73\% (330) without adding noise features. However, the duration also increases to the same duration as shapicant. \\

Other possible optimizations are also applicable to other feature selection methods, such as applying backward feature selection after \textit{powershap} to eliminate any noise features, redundant, or duplicate features. Another possibility is optimizing the used machine learning model to better match the dataset and rerun \textit{powershap}, e.g. by using more CatBoost estimators for datasets with large sample sizes and high dimensional feature sets.\\
In the benchmarking results, there are datasets where including all features perform equally well or even better, such as in the case of the Gina prior test set. In these cases, the filter methods perform well but output large feature sets, while the forward feature selection performs the worst. Alternatively, \textit{powershap} can be used here as a fast wrapper-based dimensionality reduction method to retain approximately the same performance with a much smaller feature set. As such, there will still be a trade-off for each use-case between filter and wrapper methods based on time and performance.\\
We are aware that the current design of the benchmarks has some limitations. For the simulation benchmark, the \texttt{make\_classification} function uses by default a hypercube to create its classification problem, resulting in a linear classification problem, which is inherently easier to classify~\cite{scikit-learn}. The compared filter methods were chosen by their most common usage and availability, however, these are fast and simple methods and are of a much lower complexity than \textit{powershap}. The same argument could be made for our choice of the forward feature selection method (as wrapper method) compared to other methods such as genetic algorithm based solutions. Furthermore, wrapper methods, and thus also \textit{powershap}, are highly dependent on the used model, as the feature selection quality suffers from modeling issues such as for example overfitting and underfitting. Therefore, the true potential achievable performances on the benchmark datasets may differ since every use-case and dataset requires its own tuned model to achieve optimal performance. Additionally, the cut-off values and hyperparameters of none of the methods were optimized and are either set to the same value as in \textit{powershap} or used with their default values. This might impact the performance and could have skewed the benchmark results in both directions. However, choosing the same model and the same values for hyperparameters (if possible) in all experiments, reduces potential performance differences and facilitates a fair enough comparison. 

\section{Conclusion}\label{sec:conclusion}
We proposed \textit{powershap}, a wrapper feature selection method using Shapley values and statistical tests to determine the significance of features. \textit{powershap} uses power calculations to optimize the number of required iterations in an automatic mode to realize fast, strong, and reliable feature selection. Benchmarks indicate that \textit{powershap}'s performance is significantly faster and more reliable than comparable state-of-the-art shap-based wrapper methods. \textit{Powershap} is implemented as an open-source plug-and-play \textit{sklearn} component, increasing its accessibility and ease of use, making it a power-full Shapley feature selection method, ready for your next feature set.  

\paragraph{Acknowledgements.}
Jarne Verhaeghe is funded by the Research Foundation Flanders (FWO, Ref. 1S59522N) and designed \textit{powershap}. Jeroen Van Der Donckt implemented \textit{powershap} as an \textit{sklearn} component. Sofie Van Hoecke and Femke Ongenae supervised the project. A special thanks goes to Gilles Vandewiele for proof-reading the manuscript.

\paragraph{Code.}
The code, documentation, and more benchmarks can be found using the following link: \href{https://github.com/predict-idlab/PowerSHAP}{https://github.com/predict-idlab/PowerSHAP}

\bibliographystyle{splncs04}
\bibliography{bibliography}

\end{document}